\documentclass[sigconf]{acmart}
\AtBeginDocument{%
  \providecommand\BibTeX{{%
    \normalfont B\kern-0.5em{\scshape i\kern-0.25em b}\kern-0.8em\TeX}}}

\copyrightyear{2022}
\acmYear{2022}
\setcopyright{acmcopyright}
\acmConference[SIGIR '22] {Proceedings of the 45th International ACM SIGIR Conference on Research and Development in Information Retrieval}{July 11--15, 2022}{Madrid, Spain.}
\acmBooktitle{Proceedings of the 45th International ACM SIGIR Conference on Research and Development in Information Retrieval (SIGIR '22), July 11--15, 2022, Madrid, Spain}
\acmPrice{15.00}
\acmISBN{978-1-4503-8732-3/22/07}
\acmDOI{10.1145/3477495.3531920}
\usepackage{enumitem}

\acmSubmissionID{sp2221}

\begin{document}
\fancyhead{}
\title{Task-Oriented Dialogue System as Natural Language Generation}

\author{Weizhi Wang}
\authornote{This work is done during first and third author's internship at Alibaba DAMO Academy.}
\affiliation{%
  \institution{University of California Santa Barbara}
  \city{Santa Barbara}
  \state{California}
  \country{USA}
}
\email{weizhiwang@ucsb.edu}

\author{Zhirui Zhang}
\authornote{Corresponding author and this work is done at Alibaba DAMO Academy.}
\affiliation{%
  \institution{Tencent AI Lab}
  \city{Shenzhen}
  \country{China}}
\email{zrustc11@gmail.com}

\author{Junliang Guo}
\authornotemark[1]
\affiliation{%
  \institution{Microsoft Research Asia}
  \city{Beijing}
  \country{China}}
\email{junliangguo@microsoft.com}

\author{Yinpei Dai}
\affiliation{%
  \institution{Alibaba DAMO Academy}
  \city{Hangzhou}
  \country{China}}
\email{yinpei.dyp@alibaba-inc.com}

\author{Boxing Chen}
\affiliation{%
  \institution{Alibaba DAMO Academy}
  \city{Hangzhou}
  \country{China}}
\email{boxing.cbx@alibaba-inc.com}

\author{Weihua Luo}
\affiliation{%
  \institution{Alibaba DAMO Academy}
  \city{Hangzhou}
  \country{China}}
\email{weihua.lwh@alibaba-inc.com}

\renewcommand{\shortauthors}{Wang and Zhang, et al.}


\begin{abstract}
In this paper, we propose to formulate the task-oriented dialogue system as the purely natural language generation task, so as to fully leverage the large-scale pre-trained models like GPT-2 and simplify complicated delexicalization prepossessing. 
However, directly applying this method heavily suffers from the dialogue entity inconsistency caused by the removal of delexicalized tokens, as well as the catastrophic forgetting problem of the pre-trained model during fine-tuning, leading to unsatisfactory performance.
To alleviate these problems, we design a novel GPT-Adapter-CopyNet network, which incorporates the lightweight adapter and CopyNet modules into GPT-2 to achieve better performance on transfer learning and dialogue entity generation.
Experimental results conducted on the DSTC8 Track 1 benchmark and MultiWOZ dataset demonstrate that our proposed approach significantly outperforms baseline models with a remarkable performance on automatic and human evaluations.
Source code and data are available at \url{https://github.com/Victorwz/tod_as_nlg}.

\end{abstract}

%
%






\maketitle

\section{Introduction}
The increasing use of customer service and personal assistants has spurred interest in building task-oriented dialogue systems that help users to accomplish a wide range of tasks via natural language conversations, such as weather forecast, restaurant reservation, IT helpdesk and airplane booking. 
The typical task-oriented dialogue system follows a pipeline structure that has four modules and executes sequentially.
A natural language understanding (NLU) \cite{kim2017onenet,Lee2019ConvLabME} module first identifies user intents and extracts associated information from user utterance input, based on which the dialogue state tracking (DST) module \cite{wu2019transferable, zhang2019find, le2019non, goel2019hyst, kim2020efficient} tracks the values of slots to update the belief state.
Then the dialogue policy (POL) module \cite{Peng2018IntegratingPF, Zhao2019RethinkingAS, zhang2019budgeted} decides the next system action under the belief state and database query results.
Finally, the natural language generation (NLG) module \cite{Chen2019SemanticallyCD} maps the system action to a natural language response. 
In practice, these different modules are usually optimized separately, which does not necessarily lead to overall optimized performance for task completion.

In order to overcome this drawback, one line of research attempts to integrate these pipeline modules into one single model, and builds up the end-to-end neural architecture for task-oriented dialogue systems, including Mem2Seq \cite{Madotto2018Mem2SeqEI}, Sequicity \cite{Lei2018SequicityST} and TransferTransfo \cite{budzianowski2019hello}.
Recently, researchers propose to incorporate GPT-2~\cite{radford2019language}, a large-scale pre-trained model, to construct a unified language model for task-oriented dialogue with a single sequence in format of dialogue history \cite{ham2020end,hosseini2020simple,peng2020soloist,Yang2020UBARTF},
However, these methods still maintain the traditional delexicalization prepossessing and special separators, such as ``[restaurant\_address]'' and ``<usr>'' that are appeared in the data format of Neural Pipeline GPT-2~\cite{ham2020end}, which amplifies the inconsistency between pre-training and fine-tuning stages.

\begin{figure*}[t] 
\centering 
\includegraphics[width=0.99\textwidth]{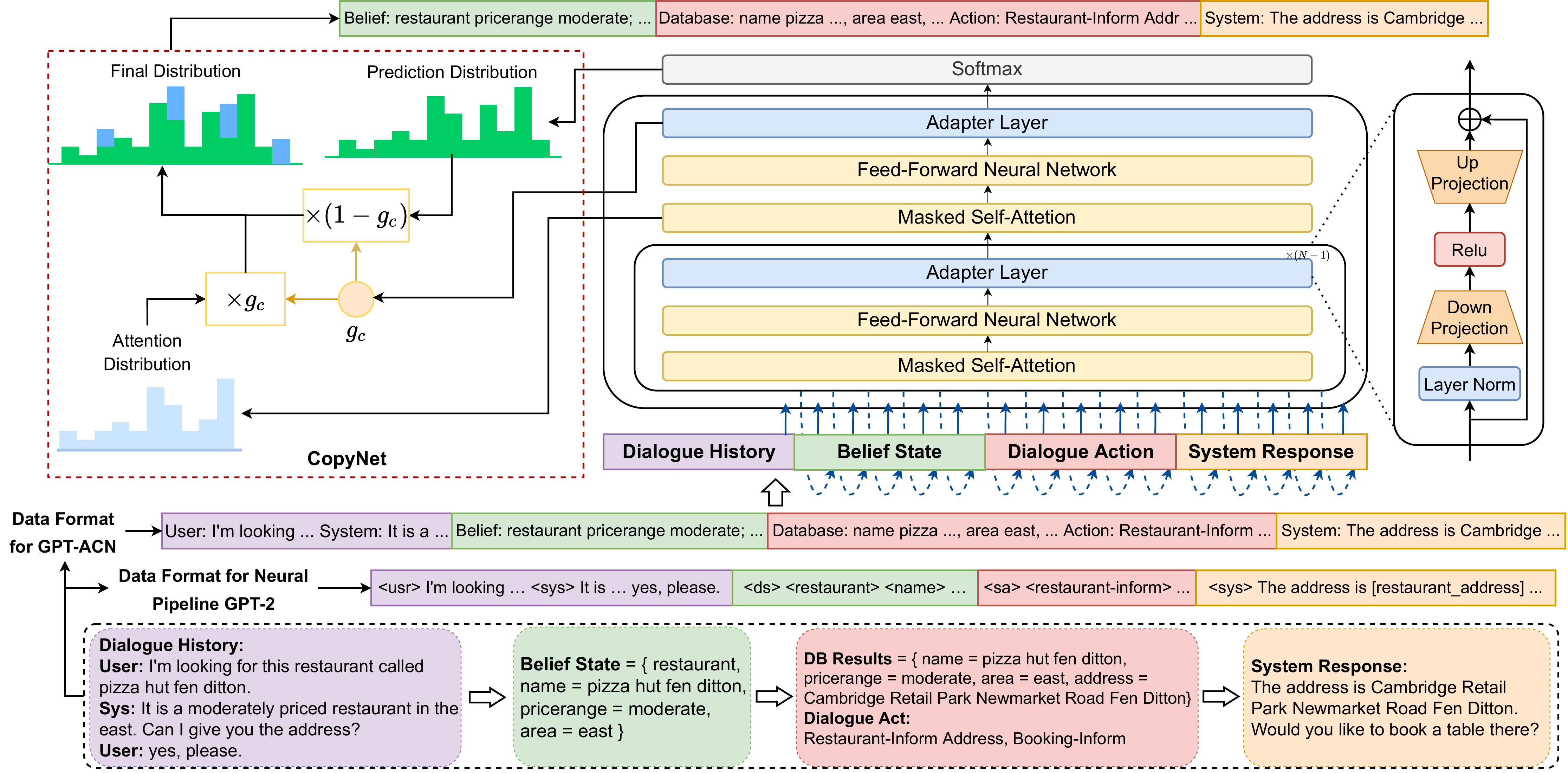} 
\vspace{-5pt}
\caption{The overview of GPT-Adapter-CopyNet network (GPT-ACN), in which database results are inserted between DST and POL modules to leverage the database information and we mask the correspondent training loss during fine-tuning.}
\label{fig:model}
\end{figure*}

On the basic of this, we propose to further convert the entire pipeline structure into a purely prefix-given NLG task by removing complicated delexicalization prepossessing and special separators, as shown in Figure \ref{fig:model}. 
In this way, instead of making pre-training more similar to a downstream task, we reformulate the task itself to make it more similar to the pre-training objective of GPT-2, which takes full advantage of language generation ability of the pre-trained model.
However, this method typically brings the dialogue entity inconsistency (e.g. names of hotels, postcodes of restaurants), which is not conducive to task completion.
It is because that the current pre-trained models have no correspondent structure to ensure entity consistency, when we directly replace the delexicalized tokens with natural text. 
On the other hand, since only a handful of examples is available for building task-oriented dialogue systems, directly fine-tuning the whole pre-trained model on this purely NLG task easily suffers from the catastrophic forgetting problem \cite{McCloskey1989CatastrophicII}, thereby degrading the pre-training performance.

In this paper, we design a novel GPT-Adapter-CopyNet network (GPT-ACN), which incorporates two simple and lightweight modules into GPT-2 during fine-tuning, to handle the aforementioned problems.
Specifically, we first introduce adapter module, which is inserted between layers of the pre-trained GPT-2 model.
When fine-tuning GPT-2 on the downstream dataset, this model only updates parameters in adapter modules and keeps pre-trained parameters frozen to alleviate the catastrophic forgetting problem, thus achieving better performance of transfer learning.
Besides, in order to ensure entity consistency in dialogue flow, we further build up a CopyNet module on the top of the GPT-2 model, which enables the model to directly copy entities from dialogue history instead of generating entities.

We evaluate the effectiveness of our proposed method on the DSTC8 Track 1 benchmark and MultiWOZ dataset.
The experimental results prove that our proposed approach significantly outperforms previous state-of-the-art baselines, achieving a remarkable performance on automatic and human evaluations.




\section{Problem Formulation}
The whole process of task-oriented dialogue system can be formulated as follows:
at turn $t$ of the dialogue, given the dialogue history with many utterances in several turns $H_{1:t} = \{U_0, S_0,  \cdots, S_{t-1}, U_t\} $, the dialogue system is supposed to obtain the belief state $B_t$, search the database results $D_t$, predict current system action $A_t$, and generate final system response $S_t$.
The conventional approach mainly splits this task into three sub-tasks, including dialogue state tracking $P(B_t |H_{1:t})$, dialogue policy prediction $P(D_t, A_t |H_{1:t}, B_t)$ and natural language generation $P(S_t|H_{1:t}, B_t, D_t, A_t )$. 
Recently, the end-to-end task-oriented dialogue systems aim to directly optimize the entire distribution $P(B_t, D_t, A_t, S_t |H_{1:t})$ with one single model~\cite{ham2020end,hosseini2020simple,peng2020soloist}.
These methods convert the dialogue state and system action to word tokens, and then concatenate the whole dialogue flow into a long sequence by delimiter tokens and additional special words, which can be learned in a self-supervised manner.
However, the additional introduced special tokens still require complicated delexicalization preprocessing and bring inconsistency between the pre-training and fine-tuning stages, which hinders the model from fully employing the capacity of the pre-trained model.
In this paper, we simplify the end-to-end task-oriented dialogue into a purely prefix-given NLG task by removing traditional delexicalization preprocessing and replacing special separators with natural text.
Since this process is independent of pipeline format, this approach could be naturally applied in all previous end-to-end task-oriented dialogue systems.
For instance, as illustrated in Figure~\ref{fig:model}, Neural Pipeline GPT-2 introduces delimiter tokens ``<usr>'', ``<sys>'', ``<ds>'' and ``<sa>'' to signal the beginning of sequence representations of user utterance, system response, dialogue state, and system action, while our proposed method directly replaces these tokens with corresponding natural text: ``User:'', ``System:'', ``Belief:'' and ``Action:''. 
Also, all delexicalized tokens such as ``[restaurant\_address]'' are replaced by its original entities.

\section{GPT-Adapter-CopyNet Network}
When converting task-oriented dialogue into a purely NLG task, it heavily suffers from the dialogue entity inconsistency caused by the removal of delexicalized tokens, as well as the catastrophic forgetting problem of the pre-trained model during fine-tuning.
To address these problems, we design a novel GPT-Adapter-CopyNet network (GPT-ACN), as shown in Figure~\ref{fig:model}. 
This whole framework uses the GPT-2 model as the backbone, which consists of a stack of transformer decoder layers.
Based on this, a simple and lightweight adapter layer is firstly injected between layers of GPT-2 transformer to alleviate the catastrophic forgetting problem.
Then a CopyNet module is built on the top of GPT-2 transformer to improve entity consistency in dialogue. 
\begin{itemize}[leftmargin=*]
\setlength{\itemsep}{0pt}
\item Adapter: We adopt the original adapter structure \cite{houlsby2019parameter,guo2020incorporating,Guo2021AdaptiveAA}. 
Each residual adapter layer first performs layer normalization towards the output hidden states of previous transformer layer.
Then it is followed by a down projection layer, a non-linear activation layer, and an up projection layer. 
Last, the residual connection \cite{he2016deep} is implemented between the input hidden states and output of up projection layer, to prevent the degradation problem in very deep model. 
Assume the output hidden states of the $i$-th layer is $h_i$, the output of adapter layer $x_{i+1}$ can be computed as:
\begin{align}
    x_{i+1} = h_i + W_{u}\cdot (\text{ReLU}(W_{d}\cdot \text{LN}(h_i))),
\end{align}
where $W_{u} \in \mathbb{R}^{A\times H}$ and $W_{d}\in \mathbb{R}^{H\times A}$ are parameters, $A$ and $H$ denote the size of adapter layer and the hidden size respectively.
\item Copy Network:
In the task-oriented dialogue, it is vital to keep some entities consistent over the dialogue flow, e.g., hotel names and restaurant postcodes.
Actually, it is easily achieved by directly copying entities from dialogue history instead of generating entities.
Based on this motivation, we introduce the CopyNet module \cite{Gu2016IncorporatingCM} on the top of the GPT-2 model, which enables the model to generate words from both copying words via pointing, and original prediction distribution. 
Specifically, at $j$-th step of model prediction, we first obtain the embedding $e_j$ of input tokens, the attention score $a^L$ of last layer, the output hidden states $h_{j}^{L}$, and the prediction probability $P_{\text{g}}(\omega)$ by the original model.
Next, the copy probability $g_{\text{c}}\in [0,1]$ is calculated by:
\begin{align}
    g_{\text{c}} = \sigma(W_{\text{c}}\cdot [e_j;h^L_j] + b_{\text{c}}),
\end{align}
where $W_{\text{c}}\in \mathbb{R}^{2\cdot H\times 1}$ and $b_{\text{c}}$ are learnable parameters, and $\sigma$ is the sigmoid function. 
The final distribution $P(\omega)$ is a soft linear combination of original probability and attention score: 
\begin{align}
    P(\omega) =(1 - g_{\text{c}} )\cdot P_{\text{g}}(\omega) + g_{\text{c}} \cdot \sum_{k:\omega_k=\omega}a_k^t.
\end{align}
\item Optimization:
The training objective of the GPT-ACN model is the standard left-to-right language modeling objective \cite{Bengio2000ANP}, which maximizes the likelihood of the next word-token from given the previous word tokens.
The GPT-ACN model does not require additional training objectives such as next-utterance classification used in \citet{ham2020end}.
For the model training, we first load pre-trained GPT-2 model checkpoint into our GPT-ACN model and then fine-tune this model with the task-oriented dialogue dataset.
During fine-tuning, only the parameters in adapter and CopyNet modules will be updated in back-propagation, keeping parameters in the original GPT-2 model fixed.
\end{itemize}

\begin{table*}[t]
\centering
\small
\begin{tabular}{l|ccccccc|cccc}
\toprule
    & \multicolumn{7}{c}{\textbf{Automatic Evaluation}}  & \multicolumn{4}{|c}{\textbf{Human Evaluation}}    \\
\midrule        
\textbf{Model}  & \textbf{Succ.$\uparrow$}  & \textbf{Book.$\uparrow$}   & \textbf{Return$\uparrow$}              & \textbf{Turns$\downarrow$} & \textbf{Prec.$\uparrow$}   & \textbf{Recall$\uparrow$} &\textbf{F1$\uparrow$}           & \textbf{Succ.$\uparrow$}    & \textbf{Under.$\uparrow$} & \textbf{Appr.$\uparrow$} & \textbf{Turns$\downarrow$}                \\
\midrule
\midrule
ConvLab Baseline  & 62.00\% & 84.38\% & 30.41 & 7.67 & 0.72 & 0.83 & 0.75  & 57.0\% & 3.10 & 3.56 & 17.54   \\
NP-GPT            & 78.60\% & 86.34\% & 48.92 & \textbf{7.40} & \textbf{0.87} & 0.89 & \textbf{0.87} & 69.0\% & 4.02 & 4.46 & 16.52  \\
NP-NLG            & 78.00\% & 81.33\% & 47.52 & 8.08 & 0.78 & 0.92 & 0.83 & 72.0\% & 4.10 & 4.64 & 17.02 \\
NP-GPT-ACN (Ours) & \textbf{82.80\%}  & \textbf{90.97\%} & \textbf{53.36} & 8.00 & 0.79 & \textbf{0.95} & 0.84 & \textbf{76.0\%} & \textbf{4.32} & \textbf{4.72} & \textbf{15.44} \\
\midrule
\midrule
- w/o NLG         & 80.00\% & 89.25\% & 50.01 & 7.99 & 0.85 & 0.90  & 0.86  & - & - & - & -  \\
- w/o CopyNet     & 80.20\% & 85.45\% & 50.76 & 7.68 & 0.76 & 0.94 & 0.82  &  - & - & - & -  \\
\bottomrule
\end{tabular}
\caption{\label{tab:conv}Model performance on ConvLab automatic and human evaluation. \textit{Succ.}, \textit{Book.}, \textit{Prec.}, \textit{Under.} and \textit{Appr.} are short for success rate, book rate, precision, understanding score and appropriate score. }
\vspace{-20pt}
\end{table*}

\begin{table*}[t]
\centering
\small
\begin{tabular}{l|c|ccccc|cccc}
\toprule
      &            & \multicolumn{5}{c}{ \textbf{DST and Context-to-Text Generation}} & \multicolumn{4}{|c}{\textbf{End-to-End Response Generation}} \\
\midrule
\textbf{Model} & \textbf{Extra.} & \textbf{Joint Accuracy} & \textbf{Inform} & \textbf{Success} & \textbf{BLEU} & \textbf{Combined}  & \textbf{Inform} & \textbf{Success} & \textbf{BLEU} & \textbf{Combined}    \\
\midrule
KAGE-GPT2~\cite{Lin2021KnowledgeAwareGG}&  $\times$  & 54.86 & - & - & - & - &  - & - & - & -  \\
DAMD~\cite{zhang2020task} &\checkmark  &  - & 89.20 & 77.90 & 18.60 & 102.15 & 76.40 & 60.40 & 16.60 & 85.00      \\
SOLOIST~\cite{peng2020soloist} & \checkmark & - & 89.60 & 79.30 & 18.03 & 102.49 & 85.50 & 72.90 & 16.54 & 95.74     \\
\midrule
\midrule
SimpleTOD~\cite{hosseini2020simple} & $\times$ & 50.22\textsuperscript{$\dagger$} & 88.90 & 67.10 & 16.90 & 94.90 &  84.40 & 70.10 & 15.01 & 92.26   \\
SimpleTOD-NLG      & $\times$ & \textbf{56.23} & 84.80 & 70.30 & 14.88 & 92.43 & 79.60 & 63.80 & 14.23 & 85.93   \\
SimpleTOD-GPT-ACN (Ours)       & $\times$ & 55.57 & \textbf{93.70} &  \textbf{76.70} &  \textbf{17.02} &  \textbf{102.22}    &  \textbf{85.80} & \textbf{72.10} & \textbf{15.52} & \textbf{94.47} \\
\bottomrule
\end{tabular}
\caption{\label{tab:multiwoz}Model performance on MultiWOZ 2.0 Dialogue State Tracking (DST), Context-to-Text Generation and End-to-End Response Generation. The result with mark ($\dagger$) is our reproduction due to the lack of result in \citet{hosseini2020simple}. Checking for ``Extra." means that the model adds extra pre-trained data or performs data augmentation.}
\vspace{-10pt}
\end{table*}

\section{Experiments}

\subsection{Setup}
\paragraph{Datasets and Metrics.} 
We evaluate the effectiveness of our approach on two task-oriented dialogue benchmarks: DSTC8 Track 1 End-to-End Multi-Domain Dialogue Challenge~\cite{kim2019eighth} and MultiWOZ 2.0 benchmark with three sub-tasks ~\cite{Budzianowski2018MultiWOZA}.
The automatic and human evaluations of DSTC8 Track 1 is carried out by ConvLab~\cite{Lee2019ConvLabME}, an open-source multi-domain end-to-end dialogue system platform: 
\begin{itemize}[leftmargin=*]
\setlength{\itemsep}{0pt}
\item Automatic evaluation with user simulator: Success Rate, Book Rate, Return, Turns, Precision, Recall, F1. As for the \textit{Success Rate}, the dialogue is considered as successful only if the requestable slots are correctly filled and book success if needed.
The book success is achieved only if the reserved information fits into all informable slots, and it is considered as a sub-evaluation called \textit{Book Rate}.
Also, \textit{Precision}, \textit{Recall}, and \textit{F1} measure the accuracy of requestable slot filling.
\textit{Return} is a reward score obtained from the user simulator when the task is finished and we follow the same calculation method as \citet{ham2020end}. 
The maximum limit of turns in one dialogue is set to 40 in our experiments.
\item Human evaluation with crowd-workers: Success Rate, Language Understanding Score, Response Appropriateness Score, Turns. \textit{Language Understanding Score} and \textit{Response Appropriateness Score} are the metrics of how natural the response of the model is, with the 5 point scale. 
\end{itemize}
We follow the automatic evaluation metrics to evaluate task completion and response quality for MultiWOZ 2.0 benchmark: 
\textit{Inform} measures whether a system has provided a correct entity, \textit{Success} verifies whether it has answered all the requested information, and \textit{BLEU} \cite{Papineni2002BleuAM} is used to measure the fluency of the generated responses.
A combined score (\textit{Combined}) is also reported as an overall quality measure suggested in \citet{Mehri2019StructuredFN}, which is computed with \textit{(Inform +Success)×0.5+BLEU}. 
Besides, \textit{Joint Accuracy} is adopted to evaluate dialogue state tracking task.

\paragraph{\label{sec:baselines}Baselines and Details.}
We compare GPT-ACN with two strong baselines based on GPT-2: (\textit{i}) Neural Pipeline GPT-2 (NP-GPT) \cite{ham2020end} integrates the pipeline modules into one single pre-trained auto-regressive language model and realizes end-to-end training and inference. 
This model wins the DSTC8 Track 1 End-to-End Multi-Domain Dialogue Challenge with the No.1 performance on human evaluation; (\textit{ii}) SimpleTOD \cite{hosseini2020simple}, similar to NP-GPT, also simplifies pipeline modules to a long sequence, and use a self-defined format for dialogue pipeline.
It is considered as our baseline for MultiWOZ 2.0 dataset.
We implement the purely natural text version of NP-GPT and SimpleTOD by removing delexicalization preprocessing and replacing special separators with natural text.
These two baselines are named as NP-NLG and SimpleTOD-NLG respectively, while we apply the GPT-ACN model for these baselines and construct the corresponding version of our approach, called NP-GPT-ACN and SimpleTOD-GPT-ACN.
As our method skips the delexicalization stage, we insert up to three database query results between DST and POL modules in these four systems to leverage the database information.
All models are developed with HuggingFace’s Transformers \cite{Wolf2019HuggingFacesTS} and adopt \texttt{GPT-2-small} model ($n_{\text{layer}}=12,n_{\text{head}}=12, d_{\text{embd}}=768$) with 117 million parameters as backbone.
The maximum number of turns in dialogue history is set to 15 and the size of adapter layer is set to $512$.
The Adam \cite{kingma2014adam} optimizer with a learning rate of 3e-4 is used in our experiments and we train all models with a batch size of 2 for 15 epochs.
During inference, we adopt the simplest greedy decoding.
Since our approach directly generates the natural text responses, we apply the delexicalization processing for the model output in MultiWOZ 2.0 benchmark, making it comparable to previous baselines.  

\subsection{Performance on DSTC8 Track 1 }
The automatic and human evaluation results are listed in Table~\ref{tab:conv}.
For automatic evaluation, following \citet{ham2020end}, all systems are required to accomplish 500 dialogue tasks with pre-defined user goal and the same environment. 
We can find the performance degradation on precision and book rate between NP-GPT and its natural text version NP-NLG.
It shows that simplifying end-to-end task-oriented dialogue as purely NLG makes it a harder task, in which the removal of delexicalization brings difficulty in entity generation consistency.
Thus, directly fine-tuning GPT-2 cannot make accurate generation on requestable slots and values.
Instead, the introduction of our GPT-ACN architecture solves this problem well, in which NP-GPT-ACN exceeds NP-GPT by 4.20\% success rate, 4.63\% book rate, and 4.44 return. 
Besides, removing NLG leads to significant performance degradation of NP-GPT-ACN, which indicates that converting into a purely NLG task will bring more improvement space for fully exploiting the pre-trained model.

For the human evaluation on ConvLab, the task-oriented dialogue systems are evaluated in Amazon Mechanic Turk, in which crowd-workers communicate with systems and provide a rating based on the whole experience. 
Every model is evaluated on 100 tasks sampled from the user goal list.
We can see that NP-GPT-ACN significantly outperforms NP-GPT with a success rate of 76.0\%, a language understanding score of 4.32, a response appropriateness score of 4.72. 
These results validate the effectiveness of our method on fully exploiting the pre-trained model.
Surprisingly, compared with NP-GPT, NP-NLG achieves better performance on success rate and response appropriateness score, which indicates that NP-NLG could produce more natural responses.

\begin{figure}[t] 
\centering 
\includegraphics[width=0.45\textwidth]{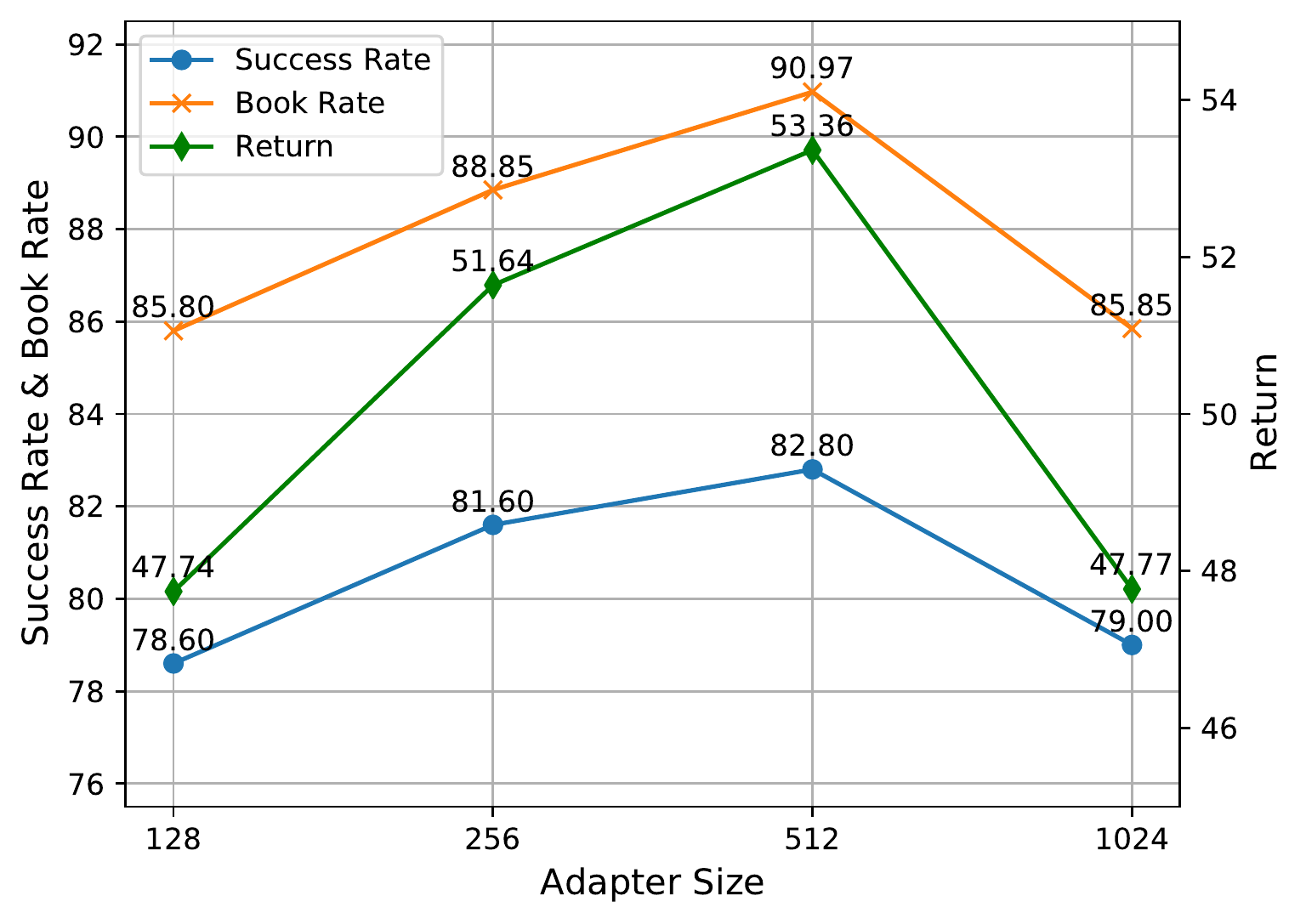} 
\vspace{-10pt}
\caption{The effect of adapter bottleneck size on ConvLab automatic evaluation. The selected metrics are success rate, return and book rate.} 
\label{fig:adapter}
\end{figure}


We further study the impact of the adapter and CopyNet modules.
As shown in the last row of Table~\ref{tab:conv}, the NP-GPT-ACN suffers from an obvious loss on precision and book rate, when the CopyNet is removed from model architecture.
It proves that incorporating CopyNet enables the model to directly copy entities, thus achieving better entity consistency in dialogue flow.
The ``- w/o CopyNet'' still yields better performance then NP-NLG thanks to the introduction of the adapter module.
It verifies that keeping the original GPT-2 model fixed and only fine-tuning additional parameters can better alleviate the catastrophic forgetting problem.
Since the adapter bottleneck size is the major hyper-parameter in the NP-GPT-ACN model, we carry out another ablation study to evaluate its effectiveness with different sizes $\{128,256,512,1024\}$, as illustrated in Figure~\ref{fig:adapter}. We can observe that the model with an adapter layer size of 512 performs best in our experiments.


\subsection{Performance on MultiWOZ Benchmark }
We conduct the end-to-end training for our method with the MultiWOZ 2.0 dataset and directly evaluate the performance on three sub-tasks. The experimental results as shown in Table~\ref{tab:multiwoz}:
\begin{itemize}[leftmargin=*]
\setlength{\itemsep}{0pt}
\item Dialogue State Tracking: We reproduce the SimpleTOD model with the joint accuracy of 50.22 and there is a big performance degradation between SimpleTOD and SimpleTOD-NLG.
The delexicalization preprocessing leads to worse performance, which is also observed by \citet{Yang2020UBARTF}.
Besides, we compare our method with the state-of-the-art baseline KAGE-GPT2~\cite{Lin2021KnowledgeAwareGG}. 
Both SimpleTOD-NLG and SimpleTOD-GPT-ACN outperform this baseline, where SimpleTOD-NLG achieves state-of-the-art performance with a 56.23 score.
\item Context-to-Text Generation: 
This sub-task requires the system to generate actions and responses with ground truth belief states and database results.
SimpleTOD-GPT-ACN significantly outperforms two baselines SimpleTOD and SimpleTOD-NLG in all metrics, which verifies the effectiveness of our method.
In addition to SimpleTOD, our method achieves comparable performance with DAMD and SOLOIST, which add extra pre-trained data and perform data augmentation for performance improvement.
SimpleTOD-GPT-ACN actually could achieve better performance through these strategies and we leave it as future work. 
\item End-to-End Modeling:
The dialogue systems are supposed to generate belief state, dialogue action and system response sequentially. 
SimpleTOD-GPT-ACN still gains significant improvements compared with SimpleTOD and SimpleTOD-NLG, same as the previous sub-task.
The performance on three sub-tasks also demonstrates the simplicity of our proposed method that optimizes all pipeline modules jointly. 
\end{itemize}

\section{Conclusion and future work}  
In this paper, we propose to simplify the end-to-end task-oriented dialogue system as a purely natural language generation task to alleviate the inconsistency of the pre-trained model between the pre-training and fine-tuning stages.
Based on this, we further incorporate two simple and lightweight modules, adapter and CopyNet modules, into the GPT-2 model to achieve better performance on transfer learning and dialogue entity generation.
Experiments conducted on the DSTC8 Track 1 and MultiWOZ 2.0 benchmark demonstrate that our proposed method can fully exploit the pre-trained model and achieve significant improvements over state-of-the-art baselines.
In the future, one interesting direction is to explore the performance of in-domain incremental pre-training with our model as suggested in \citet{Gururangan2020DontSP}.
We also would like to unify task-oriented and chit-chat dialogues with our proposed model.

\section*{Acknowledgments}
We would like to thank the anonymous reviewers for the helpful comments. This work is supported by Alibaba Innovative Research Program. We appreciate Jian Sun, Yongbin Li, and Luo Si for the fruitful discussions. 







\bibliographystyle{ACM-Reference-Format}
\bibliography{ref}










\end{document}